\documentclass[letterpaper, 10 pt, conference]{ieeeconf}  

\IEEEoverridecommandlockouts                              

\usepackage[font=footnotesize]{caption}
\usepackage{graphics} 
\usepackage{amsmath} 
\usepackage{amssymb}  
\usepackage{booktabs}
\let\labelindent\relax
\usepackage{enumitem}

\usepackage{makecell} 
\usepackage{mathtools} 
\usepackage{multirow} 

\usepackage{subfig}
\usepackage{tabularx} 
\usepackage{array}

\newcommand{\subf}[2]{%
  {\small\begin{tabular}[t]{@{}c@{}}
   \mbox{}\\[-\ht\strutbox]
   #1\\#2
   \end{tabular}}%
}

\newcolumntype{?}{!{\vrule width 1.5pt}} 

\usepackage[dvipsnames]{xcolor} 
\definecolor{mygreen}{RGB}{0, 170, 0}
\definecolor{green_rgb_56_87_35}{RGB}{56, 87, 35}

\usepackage{tikz}

\newcommand*\circled[1]{\tikz[baseline=(char.base)]{
            \node[shape=circle,color=black, draw,inner sep=0.8pt, minimum size=2pt] (char) {#1};}}
\newcommand{\rpoint}[1]{\circled{{\fontfamily{pcr}\selectfont\footnotesize{#1}}}}

\usepackage{url}
\usepackage{comment}

\usepackage[capitalize]{cleveref}
\crefname{section}{Sec.}{Secs.}
\Crefname{section}{Section}{Sections}
\Crefname{table}{Table}{Tables}
\crefname{table}{Tab.}{Tabs.}

\title{\LARGE \bf
LaneSNNs: Spiking Neural Networks for Lane Detection\\
on the Loihi Neuromorphic Processor\\
\vspace*{-10pt}
}

\author{Alberto Viale$^{1,2,*}$\thanks{*These authors contributed equally to this work.}, Alberto Marchisio$^{1,*}$, Maurizio Martina$^2$, Guido Masera$^2$, Muhammad Shafique$^3$\\
\textit{$^1$Technische Universit{\"a}t Wien, Austria}\ \ \ \textit{$^2$Politecnico di Torino, Italy}\ \ \ \textit{$^3$New York University Abu Dhabi, UAE}\\
\small{\textit{Email: \{alberto.viale, alberto.marchisio\}@tuwien.ac.at, \{maurizio.martina, guido.masera\}@polito.it, muhammad.shafique@nyu.edu}
}\\
\vspace*{-25pt}
}

\usepackage{fancyhdr,lipsum}
\fancypagestyle{firstpage}{
  \fancyhf{}
  \fancyhead[C]{To appear at the 2022 IEEE/RSJ International Conference on Intelligent Robots and Systems (IROS 2022).}
  \fancyfoot[C]{\thepage}
}

\begin{document}

\maketitle
\thispagestyle{empty}
\pagestyle{empty}
\thispagestyle{firstpage}

\begin{abstract}

Autonomous Driving (AD) related features represent important elements for the next generation of mobile robots and autonomous vehicles focused on increasingly intelligent, autonomous, and interconnected systems. The applications involving the use of these features must provide, by definition, real-time decisions, and this property is key to avoid catastrophic accidents. Moreover, all the decision processes must require low power consumption, to increase the lifetime and autonomy of battery-driven systems. These challenges can be addressed through efficient implementations of Spiking Neural Networks (SNNs) on Neuromorphic Chips and the use of event-based cameras instead of traditional frame-based cameras.

In this paper, we present a new SNN-based approach, called \textit{LaneSNN}, for detecting the lanes marked on the streets using the event-based camera input. We develop four novel SNN models characterized by low complexity and fast response, and train them using an offline supervised learning rule. Afterward, we implement and map the learned SNNs models onto the Intel Loihi Neuromorphic Research Chip. For the loss function, we develop a novel method based on the linear composition of Weighted binary Cross Entropy (WCE) and Mean Squared Error (MSE) measures. Our experimental results show a maximum Intersection over Union (IoU) measure of about 0.62 and very low power consumption of about 1 W. The best IoU is achieved with an SNN implementation that occupies only 36 neurocores on the Loihi processor while providing a low latency of less than 8 ms to recognize an image, thereby enabling real-time performance. The IoU measures provided by our networks are comparable with the state-of-the-art, but at a much low power consumption of 1 W. 

\end{abstract}


\section{Introduction}

In recent years, the design of reliable and efficient Autonomous Driving (AD) systems has become one of the key research directions of the incoming Smart mobility~\cite{Smart_mobility}. Therefore, it leads to the development of increasingly advanced algorithms and solutions. This paper proposes a class of Spiking Neural Networks (SNNs) that are directly implementable on one of the most advanced neuromorphic hardware for energy-efficient real-time deployment of advanced AD systems. Furthermore, we leverage the event-based cameras as the vision sensor, due to their appealing properties, such as energy-efficiency, biological plausibility, high dynamic range, and good compatibility with neuromorphic systems. 

\subsection{Target Research Problem and Research Challenges}

To be able to drive safely, mobile robots and autonomous vehicles must continuously analyze the surrounding environment and must take into account any slightest variation to make the best decision and to prevent catastrophic accidents. 
Hence, it is essential that the \textit{decision process} takes place in real time. Moreover, it is desirable that the developed AD system maintains low energy consumption\footnote{Note that high-performance GPUs, generally used to face Artificial Intelligence problems, take high power and area, and generate heat (requiring big coolants and package) which makes them infeasible to be placed in the electronic control units (ECUs).}, especially with its placement into battery-driven electric means of transport.

To better analyze a general AD problem, we can divide the decision process into two parts, which must follow the low-latency and low-power constraints:

\begin{enumerate}[leftmargin=*]
    \item \textbf{Vision}: the external environment is evaluated and captured by one or more sensors;
    \item \textbf{Computation}: the sensed data is analyzed and the essential information to predict the reaction of the system is provided.
\end{enumerate}

The vision system can be represented by cameras collecting images of the environment. The advanced dynamic vision sensors (DVS) enable event-based cameras that are specialized for detecting illumination changes, which mimic the behavior of the retina. They are very reactive, robust, and low-power devices~\cite{Marchisio2021RSNN}. Therefore, they represent an efficient choice for advanced AD applications~\cite{viale2021carsnn}. 


A modern trend to address the complex AD problems is to deploy \textit{Deep Neural Networks (DNNs)} that achieve high performance, but they are very expensive in terms of power consumption~\cite{Capra2020SurveyDNN}. 
An alternate trend is to leverage the emerging \textit{Spiking Neural Networks (SNNs)}~\cite{Putra2022TinySNN}. Compared to DNNs, these SNNs have higher biological plausibility and exhibit event-based processing, thereby rendering them as a low-latency and energy-efficient choice for AD tasks. 
To further achieve low power consumption and low latency, the Neuromorphic Chips provide excellent hardware platform options~\cite{Schuman2017SurveyNeuromorphic,Massa2020EfficientSNN}. 
In this paper, we focus on the \textit{``Lane detection''} problem. 

Following these research targets, we design, optimize, and implement SNNs on the Intel Loihi Neuromorphic Research Chip~\cite{Loihi_chip}, and evaluate them on the DET dataset~\cite{det}. Moreover, the vision system is based on a DVS event-based camera~\cite{Gallego_2020,Celex_V_DVS}.

\subsection{Our Novel Contributions}

\begin{figure}[h]
    \centering
    \includegraphics[width=\linewidth]{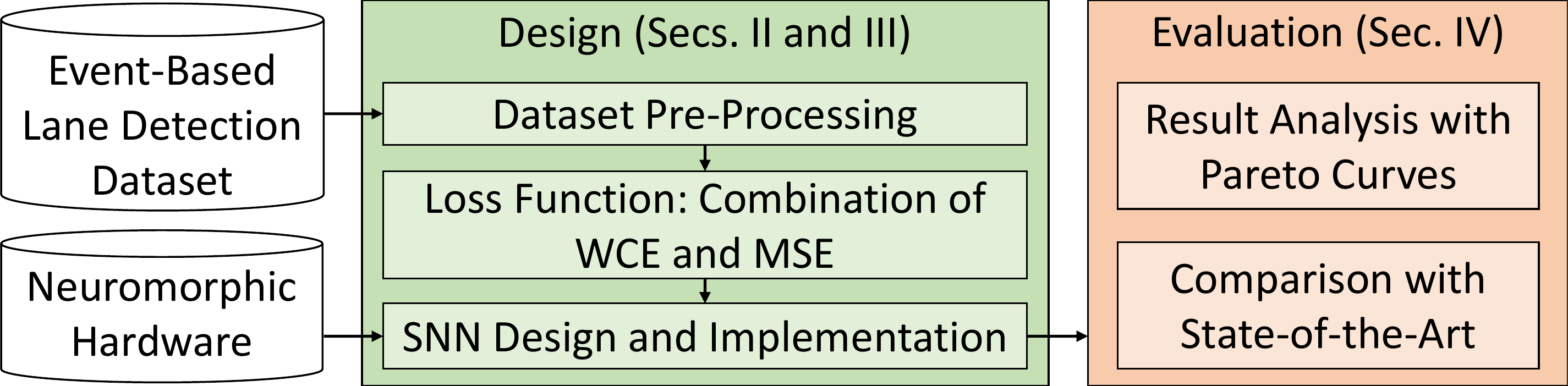}
    \caption{Overview of our novel contributions.}
    \label{fig:overview_novel_contributions}
    \vspace{-.5em}
\end{figure}

We introduce \textbf{LaneSNNs} to detect pixels that represent the lanes on general images collected by an event-based camera. An overview of our novel contributions is shown in \cref{fig:overview_novel_contributions}.
In particular, our key contributions are:

\begin{itemize}[leftmargin=*]
    \item we follow the \textit{Semantic Segmentation} approach to implement the algorithms (\textbf{\cref{cap_LaneSNN:Analysis_and_general_decisions:Lane_detection_methods}});
    \item we adopt a \textit{dataset pre-processing unit} to reduce the resolution of input and output images and to guarantee low complexity (\textbf{\cref{cap_LaneSNN:Analysis_and_general_decisions:Dataset_pre-processing}});
    \item we introduce a \textit{novel loss function} that provides a trade-off between the \textit{Weighted Binary Cross Entropy} and the \textit{Mean Squared Error} measures (\textbf{\cref{cap_LaneSNN:Analysis_and_general_decisions:Learning_rule_and_loss_function}});
    \item we implement the SNNs on the \textit{Intel Loihi Neuromorphic Research Chip}~\cite{Loihi_chip} (\textbf{\cref{cap_LaneSNN:Evaluation_of_LaneSNN:LaneSNNs_Online_implementation}});
    
\end{itemize}

As evaluation, we analyze results in form of different \textit{Pareto Curves}~\cite{Pareto_Curves} (\textbf{\cref{cap_LaneSNN:Evaluation_of_LaneSNN:Pareto_optimal_solutions}}) and we compare our results with the state-of-the-art (\textbf{\cref{cap_LaneSNN:Evaluation_of_LaneSNN:comparison_state_of_the_art}}).


\textbf{Paper Organization:} \textbf{\cref{cap_LaneSNN:Analysis_and_general_decisions}} presents our target research problem and the general design decisions. \textbf{\cref{cap_LaneSNN:LaneSNNs_design_and_overfitting_strategies}} discusses the LaneSNNs design and the anti-overfitting strategies. \textbf{\cref{cap_LaneSNN:Evaluation_of_LaneSNN}} evaluates the experimental results, and the implementations of our LaneSNNs onto the Loihi chip. \textbf{\cref{sec:conclusion}} concludes the paper.

\section{Problem Analysis and General Decisions}
\label{cap_LaneSNN:Analysis_and_general_decisions}


\subsection{Lane Detection Methods}
\label{cap_LaneSNN:Analysis_and_general_decisions:Lane_detection_methods}

\begin{figure}[h]
    \centering
    \includegraphics[width=\linewidth]{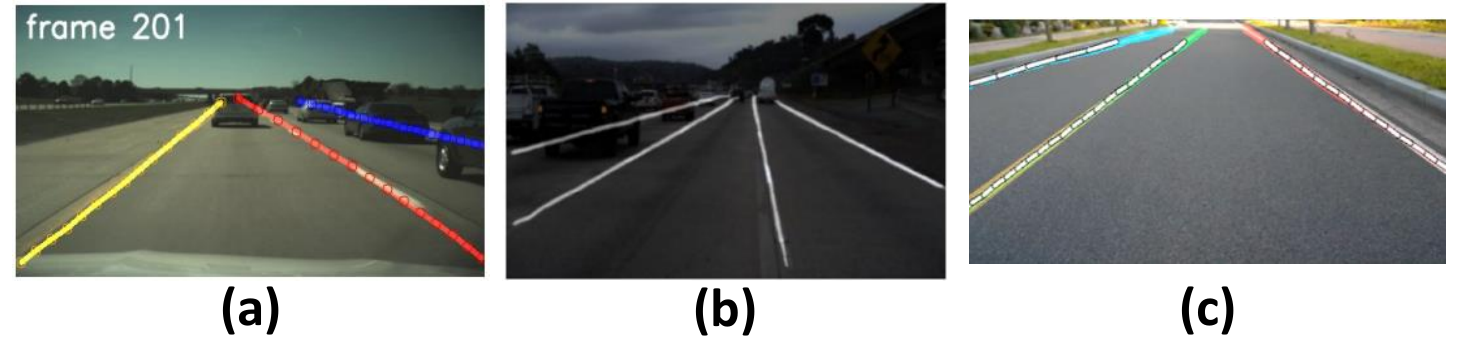}
    \caption{Example of the (a) Object Detection~\cite{regression_lane}, (b) Semantic Segmentation~\cite{s21020400} and (c) Instance Segmentation~\cite{electronics9010158} approaches for the lane detection problem.}
    \label{fig:example_of_lane_detection_methods}
    \vspace{-1.2em}
\end{figure}

The lane detection problem is one of the key tasks in the AD field. Our goal is to design and develop a device that automatically recognizes which parts of an image collected by a camera represent the lanes marked on the street. 
In the literature there exist three general classes of methods used to detect and recognize sub-parts of an image~\cite{TANG2021107623}:

\begin{enumerate}[leftmargin=*]
    \item \textbf{Object detection} (\cref{fig:example_of_lane_detection_methods} (a)): the device recognizes the coordinates of some points which constitute the lanes~\cite{regression_lane}. After that, to have an output image, these results must be post-processed to obtain the labeled image, thus increasing latency and power consumption.
    \item \textbf{Semantic segmentation} (\cref{fig:example_of_lane_detection_methods} (b)): the device distinguishes only two classes and finds the class of each pixel coming from the input image by looking at it individually. At the output, we can collect an image in which the pixel intensities define its class~\cite{pan2017spatial}~\cite{lee2017vpgnet}.
    \item \textbf{Instance segmentation} (\cref{fig:example_of_lane_detection_methods} (c)): it is based on the similar concepts as the semantic segmentation, but various lanes can be grouped into different classes~\cite{pan2017spatial}~\cite{lee2017vpgnet}.
\end{enumerate}

Since we need a real-time response from the detection device to leave more time for the decision-making part of the AD vehicle and we are only interested in the position of the detected lanes, we choose to use the semantic segmentation approach, which can achieve good performance with reduced latency and power consumption.

\subsection{Dataset Pre-Processing}
\label{cap_LaneSNN:Analysis_and_general_decisions:Dataset_pre-processing}

\subsubsection{Coding of Input Information into Spikes}

The DET dataset 
is made of labeled grey-scale raw images obtained through the DVS camera. To extract spiking information and directly feed the networks with them, we use the \textit{rate coding} technique. Hence, we compare pixels intensities to random values for converting them into spike trains with Poisson distribution.


\subsubsection{Reducing the Spatial Resolution}

The DET Dataset is recorded by the \textit{CeleX V} DVS camera~\cite{Celex_V_DVS}, and it is made by input and label images with high resolution in space ($1280\times800$ pixels per image for both inputs and labels). This property can be very useful during the training of AI models. In fact, it contains sufficient input information to understand how to better generalize the task. On the contrary, the labels with very high resolution induce a considerable imbalance between lane and background classes thus resulting in decreased accuracy when we use a semantic segmentation approach~\cite{regression_lane}.

Moreover, we design SNNs that can be directly implemented on the Intel Loihi Neuromorphic Chip~\cite{Loihi_chip}. The used Neuromorphic hardware has some limitations for the collecting of output spike counters related to the output neurons. The maximum number of counters depends on the map of the SNNs and the probes implemented to analyze the performance. 
Our preliminary analysis indicates that a maximum of $400$ spike counters can be implemented. Therefore, we reduce the size of label images to have only $400$ pixels. We also limit the resolution of the input image size to $1600$ pixels to be more coherent with the resized dimension of the output images.



To prevent the SNNs from a possible \textit{overfitting problem}, before reducing the image size, we perform \textit{data augmentation}. In this particular case, we use $271$ random training images and related labels, and perform on them random \textit{vertical translation} (between $-100$ and $100$ pixels) and \textit{rotation} (between $-30^\circ $ and $30^\circ$).



To reduce the size of the dataset images, we use two subsequent steps:

\begin{enumerate}[leftmargin=*]
    \item \textit{Vertical cropping}: we crop the top $300$ pixels rows and the bottom $200$ pixels rows for each image, which do not contain relevant information.
    \item \textit{Average resizing}: we resize the images from size $1280\times300$ to $80\times20$ for the inputs and $40\times10$ for the labels. This operation is made by the mechanism of area interpolation implemented through the \textit{OpenCV Python library}~\cite{itseez2014theopencv_manual}.
\end{enumerate}

For the label images, before the \textit{average resizing} step, we give the intensity value of $400$ to all the lane pixels (\textit{denormalization step}) and then, after performing the resizing operation, each pixel with intensity greater than $0$ is labeled as lane, and its value is normalized to $1$. This mechanism is necessary because we operate resizing with a large scale, and without the \textit{denormalization/normalization} step we could lose the thinnest lanes. 

\begin{figure}[h]
    \centering
    \includegraphics[width=\columnwidth]{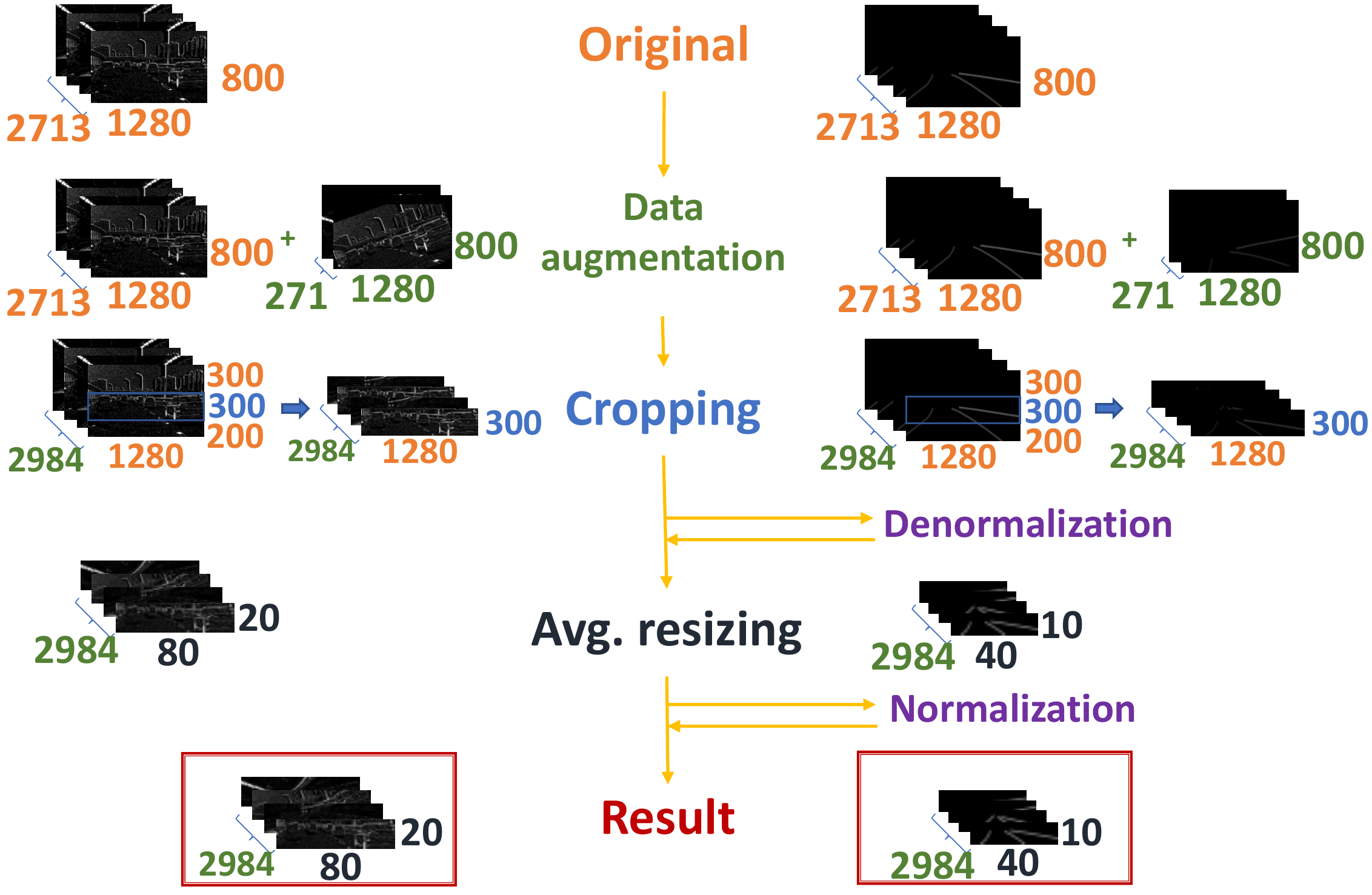}
    \caption{Steps followed to resize the images of the train set of the DET dataset. On the right, there are the input images. On the left, there are the label images. For the test set, we do not perform \textit{data augmentation}, while all the other steps remain unchanged.}
    \label{fig:dataset_manipulation}
    \vspace{-1.5em}
\end{figure}

The steps to reduce the size of the dataset images are summarized in \cref{fig:dataset_manipulation}.

\subsection{Learning Rule and Loss Function}
\label{cap_LaneSNN:Analysis_and_general_decisions:Learning_rule_and_loss_function}

\subsubsection{Learning Rule}

The DET dataset is composed of three labeled parts, namely training, validation, and testing. Therefore, it is convenient to implement a supervised learning rule for the SNNs to achieve higher performance with limited training time, rather than employing an unsupervised learning rule. In particular, we decide to use a \textit{direct supervised learning rule} to reduce the latency of the system. 
Due to the choice of obtaining the input spike trains through the \textit{rate coding} strategy (\cref{cap_LaneSNN:Analysis_and_general_decisions:Dataset_pre-processing}), the spikes are correlated in time and space, since they are based on the same image. Therefore, to achieve high performance, we use the \textit{Spatio-Temporal Back-Propagation~(STBP)}~\cite{STBP} learning rule that takes into account both temporal and spatial domains. The core of this learning rule is represented by \cref{eq:STBP_diff_eq_end_1,eq:STBP_diff_eq_end_2}. More details are discussed in~\cite{STBP}.   

\begin{equation}
\vspace{-5pt}
\frac{\partial L}{\partial \boldsymbol{b}^{n}}=\sum_{t=1}^{T} \frac{\partial L}{\partial \boldsymbol{u}^{t, \boldsymbol{n}}} \cdot \frac{\partial \boldsymbol{u}^{t, \boldsymbol{n}}}{\partial L \boldsymbol{b}^{n}}=\sum_{t=1}^{T} \frac{\partial L}{\partial \boldsymbol{u}^{t, n}}
    \label{eq:STBP_diff_eq_end_1}
\end{equation}

\begin{equation}
    \frac{\partial L}{\partial \boldsymbol{W}^{n}}=\sum_{t=1}^{T} \frac{\partial L}{\partial \boldsymbol{u}^{t, n}} \cdot \frac{\partial \boldsymbol{u}^{t, n}}{\partial x^{t, n}} \cdot \frac{\partial x^{t, n}}{\partial W^{n}}=\sum_{t=1}^{T} \frac{\partial L}{\partial \boldsymbol{u}^{t, n}} \cdot \boldsymbol{o}^{t, n-1}
    \label{eq:STBP_diff_eq_end_2}
\end{equation}

They are used to perform the \textbf{Gradient Descendent Optimization Algorithm}. 
With the implementation of the STBP learning rule, the derivative of the spiking nonlinearity is replaced by the derivative of a smooth function, following the \textbf{Surrogate Gradient} ~\cite{surrogate_model} strategy.

\subsubsection{Loss Function}

In the DET dataset, the lane and background classes are imbalanced also after the dataset pre-processing step, as described in \cref{cap_LaneSNN:Analysis_and_general_decisions:Dataset_pre-processing}. Therefore, we employ the \textit{Weighted Binary Cross-Entropy}~(WCE) loss function~\cite{Ho_2020} (\cref{eq:WCE_loss}) that is a variant of the more common \textit{Binary Cross-Entropy}~(BCE) loss function. 
\begin{equation}
L_{W C E}(y, \hat{y})=-(\beta \cdot y \cdot \log (\hat{y})+(1-y) \cdot \log (1-\hat{y})),
\label{eq:WCE_loss}
\end{equation}


where $\hat{y}$ is the predicted probability to have a lane in a determined pixel, and $y$ represents the class value that can be positive ($y=1$) or negative ($y=0$).

The WCE function introduces a little improvement for the unbalanced labels, since the positive class (i.e., the presence of the lane) gets weighted by the coefficient $\beta$ that balances the positive and negative prediction.

However, the STBP learning rule~\cite{STBP} is always studied with the implementation of the \textit{Mean Squared Error}~(MSE) loss function (\cref{eq:MSE_loss}), which is not widely used for segmentation problems~\cite{Jadon_2020}.
\begin{equation}
L_{M S E}=\frac{\sum_{i=1}^{n}(y_{i}-\hat{y}_{i})^{2}}{n}
\label{eq:MSE_loss}
\end{equation}

Therefore, for the lane detection task, we develop a novel loss function that can combine the benefits of both MSE and WCE. Such a joint weighted loss function can be formalized by the \cref{eq:MSE_BCE_loss}.
\begin{equation}
L_{MSE\ \&\ WCE}=(1-p) \cdot L_{M S E} + p \cdot L_{W C E},
\label{eq:MSE_BCE_loss}
\end{equation}

where $p$ denotes the parameter to weight the contribution of WCE and $(1-p)$ denotes the contribution of MSE, such that $0 \geq p \geq 1$. 

\section{LaneSNNs Design and Strategies to avoid Overfitting}
\label{cap_LaneSNN:LaneSNNs_design_and_overfitting_strategies}

\begin{figure}[h]
    \centering
    \includegraphics[width=7cm]{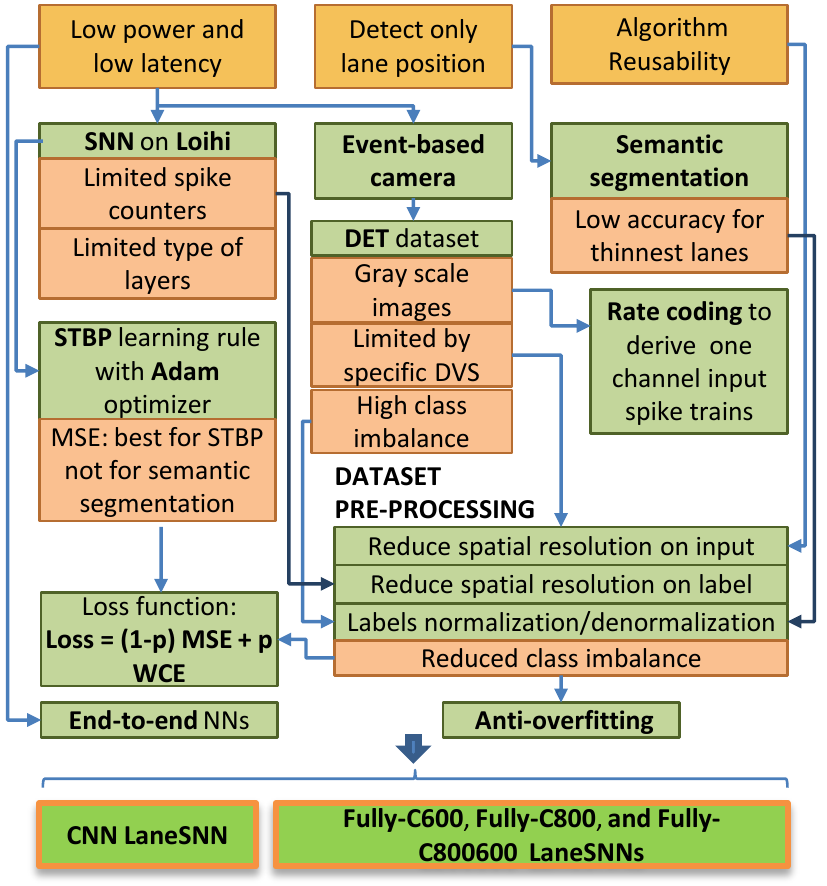}
    \caption{Design methodology of \textit{LaneSNNs} models. On the top, there are the main three desired properties (orange boxes). In the middle, the different \textit{design steps} and \textit{decisions} (green boxes) are made to follow the properties and overcome the \textit{research challenges} (red boxes). The result is the design of four LaneSNN models at the bottom.}
    \label{fig:decision_tree_lane}
    \vspace{-1em}
\end{figure}

Based on the above discussion, we present the design of our LaneSNN networks in the following along with specific design decisions. To better summarize the most important steps of our design, we present our \textit{design methodology} in \cref{fig:decision_tree_lane}.

\subsection{Input and Output}
\label{cap_LaneSNN:LaneSNNs_design_and_training_offline:LaneSNNs models design}

To be coherent with the decisions made on the DET dataset discussed in \cref{cap_LaneSNN:Analysis_and_general_decisions:Dataset_pre-processing}, since we generate only one spike-train per pixel from each raw gray-scale image, we develop our networks with only one input channel.

The output size must be consistent with the size of the label images of the modified DET dataset. Therefore, the last layer should have 400 output neurons, one for each pixel. Their firing rates represent the probability for the related pixel to be a lane in the resulting image.

\subsection{Network Architecture}

In literature, there are many examples of algorithms based on NNs for facing general semantic segmentation problems. They can be classified into two different classes according to their implementation:

\begin{itemize}[leftmargin=*]
    \item \textit{End-to-End}: these algorithms use only the NN without any pre and post-processing steps to face the lane detection problem. Usually, in these cases the network is divided into two subsequent parts, i.e., reducing (\textit{downsampling}) and increasing (\textit{upsampling}) the image size during the elaboration~\cite{end_to_end_classify}~\cite{munir2020ldnet}.
    \item \textit{More than one step}: the NN represents only a part of the entire detection algorithm, and it helps other more complex conventional algorithms~\cite{RANSAC}~\cite{s21020400} when these standalone NNs achieve low performance~\cite{ransac_lane}. 
\end{itemize}

To reduce the latency and power consumption of the entire system, we choose to implement \textit{End-to-End} algorithms. 
For the Loihi implementation, we choose to design the networks through the NxTF library~\cite{rueckauer2021nxtf}. It can describe only convolutional, fully-connected and average pooling layers. 

We develop a spiking CNN inspired from the analysis made by the works in~\cite{ransac_lane}~\cite{long2015fully_convolution_segmentation}. In the first work~\cite{ransac_lane} a small fully-connected network is introduced at the end of the NN as the \textit{upsampling} part. 
The second work~\cite{long2015fully_convolution_segmentation} emphasizes the importance of convolutional layers over others types for the \textit{downsampling} structure. 


Therefore, we design our first network, called \textbf{CNN LaneSNN} (see \cref{tab:network_2_steps}), 
with five convolutional layers by which the input sample image size ($80\times20$) is reduced to $20\times5$ pixels for each of the 16 channels. Then, the image enters into the upsampling part made of $400$ output neurons connected to the following convolutional layer. We adopt a \textit{dropout layer}, as discussed later in \cref{cap_LaneSNN:LaneSNNs_design_and_overfitting_strategies}. 

\begin{table}[h]
    \centering
    \caption{Structure of \textit{CNN LaneSNN}.}
    \resizebox{\columnwidth}{!}{
    \begin{tabular}{|c|c|c|c|c|c|c|}
        
        \hline Layer type& In ch. & Out ch. & Kernel size  & Padding & Stride & \% Dropout\\
        \hline
        Convolution & 1 & 4 & 3 & 1 & 1 & $-$\\
        Convolution & 4 & 4 & 3 & 1 & 1 & $-$\\
        Convolution & 4 & 8 & 3 & 1 & 2 & $-$\\
        Convolution & 8 & 8 & 3 & 1 & 1 & $-$\\
        Convolution & 8 & 16 & 3 & 1 & 2 & $-$\\
        \textit{Dropout} & 16 & 16 & $-$ & $-$ & $-$ & 10\\
        Dense & 1600 & 400 & $-$ & $-$ & $-$ & $-$\\
        \hline
    \end{tabular}
    }
    
    \label{tab:network_2_steps}
\end{table}

To further decrease the power consumption, we develop simpler structures that use only one or two hidden fully-connected layers. Moreover, low-complexity fully-connected networks are likely to consume low power and are easily implementable onto different Neuromorphic Chips. 
 
Therefore, we develop two fully connected networks with 800 and 600 neurons for the \textit{hidden layer} (\textbf{Fully-C800 LaneSNN}, \cref{tab:Fully-C800_LaneSNN}, and \textbf{Fully-C600 LaneSNN}, \cref{tab:Fully-C600_LaneSNN}, respectively), and a structure with two fully connected \textit{hidden layers} called \textbf{Fully-C800600 LaneSNN}. (\cref{tab:Fully-C800600_LaneSNN}).  

\begin{table}[h]
    \centering
    \caption{Structure of \textit{Fully-C800 LaneSNN}.}
    \begin{tabular}{|c|c|}
        \hline Layer & Number of neurons\\
        \hline
        Input & 1600 (image of size $80\times20$) \\
        Hidden & 800\\
        Output & 400 (image of size $40\times10$)\\
        \hline
    \end{tabular}
    
    \label{tab:Fully-C800_LaneSNN}
\end{table}
\vspace{-1.5em}
\begin{table}[h]
    \centering
    \caption{Structure of \textit{Fully-C600 LaneSNN}.}
    \begin{tabular}{|c|c|}
        \hline Layer & Number of neurons\\
        \hline
        Input & 1600 (image of size $80\times20$) \\
        Hidden & 600\\
        Output & 400 (image of size $40\times10$)\\
        \hline
    \end{tabular}
    \label{tab:Fully-C600_LaneSNN}
\end{table}
\vspace{-1.5em}
\begin{table}[h]
    \centering
    \caption{Structure of \textit{Fully-C800600 LaneSNN}.}
    \begin{tabular}{|c|c|}
        \hline Layer & Number of neurons\\
        \hline
        Input & 1600 (image of size $80\times20$) \\
        Hidden & 800\\
        Hidden & 600\\
        Output & 400 (image of size $40\times10$)\\
        \hline
    \end{tabular}
    \label{tab:Fully-C800600_LaneSNN}
\end{table}
\vspace{-1.2em}
\subsection{Anti-Overfitting Strategies}
\label{cap_LaneSNN:LaneSNNs_design_and_overfitting_strategies:Implemented_overfitting_strategies}

Due to the label imbalance of the DET dataset~\cite{det} discussed in \cref{cap_LaneSNN:Analysis_and_general_decisions:Dataset_pre-processing}, the networks hardly generalize the task, since they are affected by overfitting problems. 
This problem is mostly noticed in the layers of the networks that present many connections. For this reason, in the \textit{CNN LaneSNN} we insert a dropout layer between the last convolutional layer and the fully-connected layer. Since the percentage of dropout is not so high, it does not hamper the training of the structure. Its value of $10\%$ is chosen after some preliminary experiments.

On the other developed networks we do not apply dropout strategies, because, due to the small complexity of the networks, this kind of operation can drastically reduce the achieved results.

We use the \textit{Gaussian noise insertion} technique before all the layers of all the four networks. We define its entity with the \textit{relative standard deviation} $\sigma_r$. All the inserted Gaussian noise have $\sigma_r$ equal to $0.1$ for each layer of each developed network.

Finally, we apply the \textit{decoupled weight decay regularization} on every layer of every network (\cref{eq:weight_decay}~\cite{loshchilov2019decoupled}).
\begin{equation}
 w_{t+1}=(1-\lambda) \cdot w_{t}-lr \cdot \nabla f_{t}\left(w_{t}\right),
\label{eq:weight_decay}
\end{equation}

where:
\begin{itemize}[leftmargin=*]
    \item $w_{t+1}$ and $w_{t}$ are respectively the new and the old synaptic weights on which we apply the optimizer;
    \item $\lambda$ defines the rate of the weight decay per step;
    \item $\nabla f_{t}\left(w_{t}\right)$ is the $t^{th}$ batch gradient;
    \item $lr$ is the learning rate.
\end{itemize}
  


\cref{tab:overfitting_strategy} summarizes all the implemented strategies.

\begin{table}[h]
    \centering
    \caption{Implemented anti-overfitting strategies.}
    \resizebox{\columnwidth}{!}{
    
    \begin{tabular}{c|c|c|c}
        Anti-overfitting strategy & Networks & Where/when & Entity\\
        \hline
        Data augmentation & All & Train dataset & $+ 271$ images \\
        Dropout & CNN & Before output layer & $10\%$ \\
        Gaussian noise & All & Input of all layers & $\sigma_r=0.1$\\
        Weight decay & All & Optimization step & different values of $\lambda$
    \end{tabular}
    
    }
    
    \label{tab:overfitting_strategy}
    \vspace{-1em}
\end{table}

\section{Evaluation of LaneSNNs}
\label{cap_LaneSNN:Evaluation_of_LaneSNN}

As discussed in \cref{cap_LaneSNN:Analysis_and_general_decisions}, we perform the training of the network with the STBP learning rule. It uses \cref{eq:STBP_diff_eq_end_1,eq:STBP_diff_eq_end_2}~\cite{STBP} to evaluate the gradients. These computations are too complex to be executed onto the on-chip learning engine of the Intel Loihi Neuromorphic Chip. Therefore, our \textit{LaneSNNs} are trained offline and then we implement the networks achieving best results onto the neuromorphic hardware (\cref{cap_LaneSNN:Evaluation_of_LaneSNN:LaneSNNs_Online_implementation}). 

\subsection{Accuracy Definition}
\label{cap_LaneSNN:Evaluation_of_LaneSNN:Accuracy_definition}

As discussed in \cref{cap_LaneSNN:LaneSNNs_design_and_training_offline:LaneSNNs models design}, the output of our LaneSNNs represents the probability for each pixel to be a lane.

Compared to having a direct prediction of the class value, the probability prediction is a more flexible method, which allows to tune and even calibrate the threshold for how to interpret the predicted probabilities.

To derive the best threshold value for the predicted probabilities, we study the graphs that correlate the \textit{Precision} and \textit{Recall} values (\textbf{PR curves}~\cite{Pr_curves}) evaluated by \cref{eq:precision_recall}.

\vspace{-2pt}
\begin{equation}
\vspace{-2pt}
    \text { Precision }=\frac{TP}{TP+FP},\ \ \text { Recall }=\frac{TP}{TP+FN}
    \label{eq:precision_recall}
\end{equation}
\vspace{-2pt}

\begin{itemize}[leftmargin=*]
    \item \textit{Precision} is the number of lane pixel predictions matched with the label (\textit{True Positive} or \textit{TP}), divided by the number of pixels predicted as lane (\textit{True Positive} and \textit{False Positive} or \textit{FP}).
    \item \textit{Recall} is the number of lane pixel predictions matched with the label (\textit{TP}), divided by the number of lane pixels in the label (\textit{TP} and \textit{False Negative} or \textit{FN}).
\end{itemize}

Then we define the \textbf{F-measure} (\cref{eq:F-measure}) to find the best threshold to balance the two parameters.
\begin{equation}
\text { F-measure }=2\cdot \frac{ \text { Precision}\cdot \text{Recall}}{\text { Precision }+\text { Recall }}
\label{eq:F-measure}
\end{equation}

To distinguish between lanes and background classes, this parameter is computed for all the possible thresholds applied to the output probabilities, and the maximum F-measure, which corresponds to the best threshold, is selected. 
Moreover, to objectively compare the performance of our networks, we use the \textbf{Intersection over Union} (\textbf{IoU}) value (\cref{eq:J_index}~\cite{Jaccard_index}). 

\begin{equation}
IoU= \frac{|\text{Predicted lanes} \cap \text{True lanes}|}{|\text{Predicted lanes} \cup \text{True lanes}|} 
\label{eq:J_index}
\end{equation}


We calculate the best thresholds for every $N$ predicted images and we define the overall best threshold as the numerical mean of them (\cref{eq:best_th_mean}).
\begin{equation}
\overline{best\ th}= \frac{\sum\limits_{i=1}^N best\ th_i}{N} 
\label{eq:best_th_mean}
\end{equation}

Afterwards, we apply the $\overline{best\ th}$ on the results and compute the IoU value distinctly for each image (\cref{eq:IoU_from_best_th}).
\begin{equation}
IoU= \frac{\sum\limits_{i=1}^N IoU_i(\overline{best\ th})}{N} 
\label{eq:IoU_from_best_th}
\end{equation}

\subsection{LaneSNNs Experimental Setup}
\label{cap_LaneSNN:Evaluation_of_LaneSNN:LaneSNNs_Offline_Training}

\begin{figure}[h]
	\centering
	\includegraphics[width=.95\linewidth]{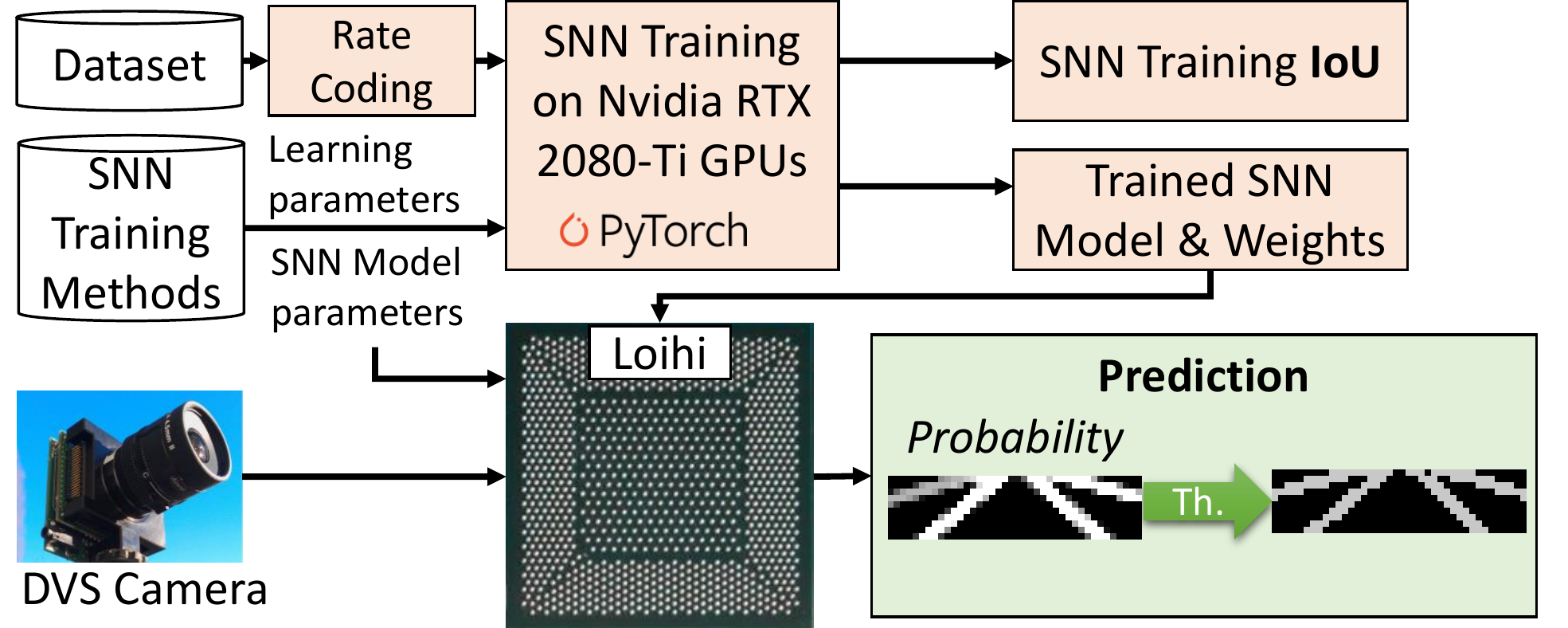}
	\caption{Setup and tool-flow for conducting our experiments.}
	\label{fig:exp_setup_DET}
	\vspace{-.5em}
\end{figure}

Our \textit{LaneSNNs}, described using the PyTorch library~\cite{PyTorch}, are trained on the DET dataset~\cite{det}, after performing the pre-processing operations discussed in \cref{cap_LaneSNN:Analysis_and_general_decisions:Dataset_pre-processing}. We run the experiments on a workstation having CentOS Linux release 7.9.2009 as the operating system and equipped with an Intel Core i9-9900X CPU and multiple Nvidia RTX 2080-Ti GPUs. An overview of the tool flow for conducting the experiments is shown in \cref{fig:exp_setup_DET}.

As discussed in \cref{cap_LaneSNN:Analysis_and_general_decisions:Learning_rule_and_loss_function}, we use the STBP~\cite{STBP} learning rule and the loss function summarized by \cref{eq:MSE_BCE_loss} for computing the distance between prediction and labels during training. 
With these modifications, the complete training set has $241837$ and $951763$ pixels representing the lane and the background classes, respectively. This corresponds to a negative (background) over positive (lanes) ratio equal to $3.93$. Therefore, to contrast the imbalance, we set the coefficient $\beta$ of \cref{eq:MSE_BCE_loss} to $4.0$ for every experiments. 
On the other hand, the value $p$, presented in the same \cref{eq:MSE_BCE_loss} and used to set the percentage of loss derived by the WCE over the MSE loss functions, varies from $0.0$ to $0.5$ for each experiment. This is because, after some experiments, we notice that the insertion of a contribution of the MSE loss function makes the SNNs converge faster. 
Focusing on other specific learning rule hyper-parameters we set:

\begin{itemize}[leftmargin=*]
    \item \textbf{Optimizer}: we use \textit{Adam}~\cite{kingma2017adam}, since it is efficient when coupled with the STBP. On that, we apply the \textit{decoupled weight decay} strategy~\cite{loshchilov2019decoupled} as discussed in \cref{cap_LaneSNN:LaneSNNs_design_and_overfitting_strategies:Implemented_overfitting_strategies}. We vary $\lambda$ of \cref{eq:weight_decay} from $0.0$ to $5e^{-4}$ with steps of $1e^{-4}$.
    \item \textbf{Learning rate} (\textbf{lr}): we use the fixed learning rate approach varying in the range from $1e^{-5}$ to $1^e{-3}$. These values are found after preliminary analyses and guarantee the convergence of the method in a few epochs. 
\end{itemize}

The adopted learning rule is directly based on the SNNs with LIF neuron models. 
The formalization of the membrane potential update ($u_i^t+1,n$) is defined in \cref{eq:STBP_model_one_formula}, where:

\begin{itemize}[leftmargin=*]
    \item $u_{i}^{t, n}$ is the membrane potential before the update;
    \vspace{2pt}
    \item $o_{i}^{t,n}$ represents the presence ($1$) or the absence ($0$) of a spike generated on the output axon;
    \vspace{2pt}
    \item $\sum_{j=1}^{l(n-1)} w_{i j}^{n} o_{j}^{t+1, n-1}$ represents the incoming synaptic weighted spikes;
    \vspace{2pt}
    \item $ b_{i}^{n}$ is a bias term.
\end{itemize}

\vspace{-15pt}
\begin{equation}
    u_{i}^{t+1, n} =u_{i}^{t, n} \tau (1-o_{i}^{t,n})+\sum_{j=1}^{l(n-1)} w_{i j}^{n} o_{j}^{t+1, n-1}+b_{i}^{n}
    \label{eq:STBP_model_one_formula}
\end{equation}

The main tunable parameters of a LIF neuron are:

\begin{itemize}[leftmargin=*]
    \item \textbf{membrane threshold}~(\textbf{$V_{th}$}): it is the same for all neurons, and its value changes from 0.2 to 1.0;
    \item \textbf{membrane reset potential}~(\textbf{$V_{reset}$}): for all the experiments, it is the same for each neuron and it is always set to $0\ V$;
    \item \textbf{membrane time constant}~(\textbf{$\tau$}): for all the experiments, it is set to $0.2\ ms$.
\end{itemize}


Moreover, the STBP learning rule uses the \textit{surrogate gradient} approach to approximate the derivative of the spiking nonlinearity with simple functions. For this purpose, we adopt the rectangular pulse function (\cref{eq:STBP_pulse_fcn}). 
\begin{equation}
   h_{1}(u)=\frac{1}{a_{1}} \operatorname{sign}\left(\left|u-V_{t h}\right|<\frac{a_{1}}{2}\right)
   \label{eq:STBP_pulse_fcn}
\end{equation}

This assumption is coherent with the work in \cite{STBP}, since different types of approximations do not involve a great variation of the accuracy, and the rectangular pulse function represents an efficient formula developed for this purpose. 

Therefore, according to \cref{eq:STBP_pulse_fcn}, we can adjust the parameter $\frac{a_1}{2}$ representing the pulse width. It is set to the same value of the $V_{th}$, as made in~\cite{STBP}.

All the experiments run for 200 epochs with batch size equal to $4$. The batch size value is set based on a preliminary analysis and it represents a trade-off between the achieved accuracy and the training time.

Based on the discussions of \cref{cap_LaneSNN:LaneSNNs_design_and_training_offline:LaneSNNs models design}, for each pixel of a gray-scale input image, we create a single spike-train. Moreover, since every single spike train is made of 30 time steps, it can contain up to 30 spikes. The spike trains per image are not calculated offline before training but are generated at run-time. Hence, they are different for each training epoch, to increase the robustness of the training process. Since this information is given as input without applying any accumulation strategy, the \textit{LaneSNNs} analyze each input image for 30 time steps. 

\subsection{LaneSNNs Implemented on Loihi}
\label{cap_LaneSNN:Evaluation_of_LaneSNN:LaneSNNs_Online_implementation}

To implement our trained LaneSNNs onto the Intel Loihi Neuromorphic Chip, we have to set its model parameters such as \textit{Compartment Voltage threshold} ($V_{th\ mant}$), \textit{Compartment Current Decay} ($\delta_i$), \textit{Compartment Voltage Decay} ($\delta_v$), \textit{Compartment Bias} ($bias$), \textit{Synaptic Weights} ($weight$) and \textit{Weight Exponent} ($wgtExp$). This is made according to the neuron models similarities exploited in~\cite{viale2021carsnn}.







We multiply weights and $V_{th}$ by a factor ($k$) calculated from the weight magnitudes of all the SNN synapses as in \cref{eq:optimum_value}:
\begin{equation}
k =\frac{2^{4}-1}{\max _{all\ synapses}(\mid  weight _{i} \mid)}
\label{eq:optimum_value}
\end{equation}

With the multiplication of the weights by $k$ we use all the dynamic range for the maximum value, thus minimizing $wgtExp$. 
All the setup parameters are summarized in \cref{tab:loihi_param_lane}.

\begin{table}[h]
    \centering
    
    \resizebox{\columnwidth}{!}{
    \begin{tabular}{c|c|c?c|c|c}
        \multicolumn{3}{c}{Offline implementation}&\multicolumn{3}{c}{Loihi implementation}
        \\
        \hline
        Parameter & Value & Precision & Parameter & Value & Precision \\ 
        \hline
        $V_{th}$ & $\times 1$ & Float 64 bits & $V_{th\ mant}$ & $\times k$ & Fixed 12 bits\\
        $weight$ & $\times 1$ & Float 64 bits & $weight$ & $\times k$ & Fixed 8 bits \\
        $\tau$ & 0.2 & Float 64 bits & $\delta_v$ & 3276 & Fixed 12 bits\\
        $b$ & 0 & Float 64 bits & $bias$ & 0 & Fixed 8 bits\\
        $-$ & $-$ & Float 64 bits & $\delta_i$ & 0 & Fixed 12 bits\\
    \end{tabular}
    }  
    \caption{Translation of parameters to the Loihi Chip for the \textit{LaneSNNs} (For the Loihi the $weight$ bits also include the $wgtExp$ bits).}  
    \label{tab:loihi_param_lane}
\end{table}

We implement all the trained LaneSNNs developed in the previous \cref{cap_LaneSNN:Evaluation_of_LaneSNN:LaneSNNs_Offline_Training} onto the Loihi Neuromorphic Chip, considering all the possible values of $V_{th}$, $\lambda$ and $lr$, characterized by the best parameter $p$. 
The implementation is conducted using the Intel Nx SDK API version 1.0.0 running onto the Nahuku32 partition. 
This code is developed using the \textit{NxTF} Layers and in particular \textit{NxConv2D} and \textit{NxDense} utilities~\cite{rueckauer2021nxtf}. 
The \textit{LaneSNNs} are tested on the testing set of the DET dataset~\cite{det}. For feeding the input images, we use the same method applied for the offline training discussed in \cref{cap_LaneSNN:Evaluation_of_LaneSNN:LaneSNNs_Offline_Training}. Hence, we create a spike train for each pixel of each input image on-the-fly. Each spike train lasts for 30 time steps. We insert a blank time of 10 time steps between two consecutive samples.

To perform the computation of the $IoU$ measure, we find the best threshold directly from the reconstructed images at the output of the Loihi chip. Then, we perform the steps discussed in \cref{cap_LaneSNN:Evaluation_of_LaneSNN:Accuracy_definition}.

\subsection{Pareto Optimal Solutions}
\label{cap_LaneSNN:Evaluation_of_LaneSNN:Pareto_optimal_solutions}

\begin{figure*}
\centering
\begin{tabular}{ccc}

\subf{\includegraphics[width=.32\linewidth]{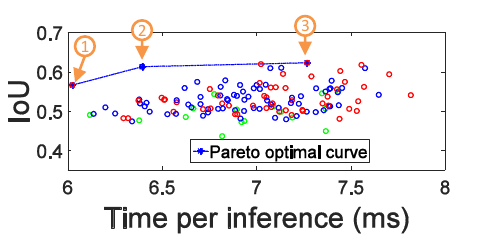}}
     {1 : $IoU$ and \textit{Latency}.}
&

\subf{\includegraphics[width=.32\linewidth]{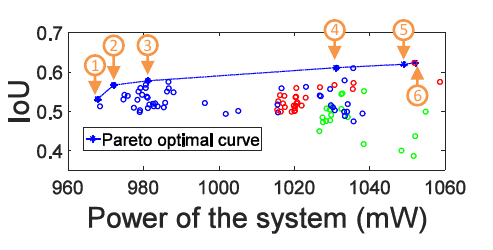}}
     {2 : $IoU$ and \textit{Power}}
&

\subf{\includegraphics[width=.32\linewidth]{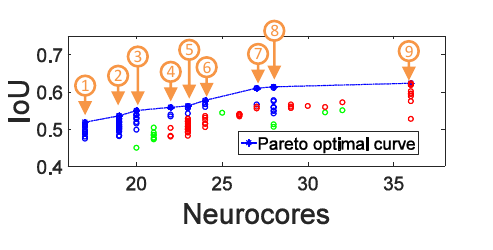}}
     {3 : $IoU$ and \textit{Complexity}}

\end{tabular}
\caption{Pareto-optimal solutions for \textcolor{red}{Fully-C800}, \textcolor{blue}{Fully-C600}, \textcolor{mygreen}{Fully-C800600} \textit{LaneSNNs}.}
\label{fig:pareto_optimal}
\vspace{-1.5em}
\end{figure*}

To efficiently evaluate the \textit{LaneSNNs} implemented onto the Intel Loihi, we can analyze multiple Pareto-optimal models. Therefore, we use the \textbf{Pareto Optimal frontier curve} 
to find the best trade-off solutions between the achieved \textbf{IoU} and:

\begin{enumerate}[leftmargin=*]
\item \textbf{latency}, i.e., the mean time duration required for the classification of all the pixels of a single image (\cref{fig:pareto_optimal}.1);
\item \textbf{power consumption} of the entire Intel Loihi chip (\cref{fig:pareto_optimal}.2);
\item \textbf{network complexity}, number of neurocores occupied (\cref{fig:pareto_optimal}.3). 
\end{enumerate}

From the first graph (\cref{fig:pareto_optimal}.1) we can see that within the Pareto-optimal curve, the maximum time to detect the lanes on a stream of 30 time steps is limited to \textbf{7.27 ms} and it is achieved by the \textit{Fully-C800} network (label~\rpoint{1}). It can be reduced to $6.02\ ms$ with a reduction on $IoU$ of about $6\%$ (label~\rpoint{3}).
In the second graph (\cref{fig:pareto_optimal}.2), we can notice that the lowest Pareto-optimal power consumption is achieved by the simplest \textit{Fully-C600} network (labels~\rpoint{1}-\rpoint{5}), but the highest $IoU$ is reached by the \textit{Fully-C800} network (label~\rpoint{6}). Overall, the power consumption of Intel Loihi chip does not vary much around $1 W$.

\cref{fig:pareto_optimal}.3 shows that, among the Pareto-optimal solutions, the \textit{Fully-C600} network is the simplest due to the lower number of occupied neurocores (labels~\rpoint{1}-\rpoint{5}). Moreover, the best $IoU$ is achieved by the \textit{Fully-C800} network, but its complexity is significantly greater than the minimum value.


\subsection{Best Results for Each LaneSNN}

The best results in terms of $IoU$ for each type of \textit{LaneSNN} for both offline and online implementations are summarized in \cref{tab:best_IoU_offline_and_online}. 

\begin{table}[h]
    \centering
    \caption{Best $IoU$ measures achieved by the different \textit{LaneSNNs} for offline (\textit{GPU}) and online (\textit{Loihi}) implementations.}
    \label{tab:best_IoU_offline_and_online}
    \resizebox{\columnwidth}{!}{
    \begin{tabular}{c|c?c|c?c|c?c|c}
        \multicolumn{2}{c}{IoU CNN}&\multicolumn{2}{c}{IoU Fully-C600}&\multicolumn{2}{c}{IoU Fully-C800}&\multicolumn{2}{c}{IoU Fully-C800600}
        \\
        \hline
        $GPU$ & $Loihi$& $GPU$ & $Loihi$& $GPU$ & $Loihi$& $GPU$ & $Loihi$\\
        \hline
         $0.598$ & $0.208$ &  $0.637$ & $0.527$ &  $0.633$ & $0.542$ &  \textbf{0.652} & $0.416$  \\
         $0.551$ & $0.349$ &  $0.632$ & \textbf{0.623} &  $0.629$ & $0.613$ &  $0.590$ & $0.550$
    \end{tabular}
    }
    \vspace*{-8pt}
\end{table}

The \textit{CNN LaneSNN} has the lowest $IoU$ values for both online and offline implementations. Its offline implementation achieves an acceptable value of $IoU$, but this result dramatically decreases for the online implementation, due to the weights approximation errors propagating layer by layer during the translation from offline to online.

The highest offline result is achieved by the \textit{Fully-C800600} network, which has two fully-connected hidden layers. However, it achieves lower $IoU$ for the online implementation than the same achieved by the simpler fully-connected networks \textit{Fully-C600} and \textit{Fully-C800}. 
The best online result is achieved by the \textit{Fully-C600} network, which is the simplest SNN. Moreover, its online and offline $IoU$ measures, even if they are slightly greater, are comparable to the results obtained with the \textit{Fully-C800} network.

\subsection{Comparison with the State-of-the-Art}
\label{cap_LaneSNN:Evaluation_of_LaneSNN:comparison_state_of_the_art}

In our work, as discussed in \cref{cap_LaneSNN:Analysis_and_general_decisions}, we prefer less complex SNNs. This means that they can be effectively implemented onto the Intel Loihi Neuromorphic Chip and achieve competitive results for real-time embedded systems, with low power consumption and low latency. This is also favored by the use of event-based cameras as vision sensors of the AD system. On the other hand, in literature, there are many algorithms that involve non-spiking NNs to face the problem of lane detection. 

For a fair comparison, we consider the results achieved by state-of-the-art networks on the same dataset (DET dataset~\cite{det}). Therefore all these NNs also use an event-based camera as the vision sensor for the AD system.


The results of the state-of-the-art methods, which are FCN~\cite{long2015fully_convolution_segmentation}, DeepLabv3~\cite{chen2017rethinking}, RefineNet~\cite{lin2016refinenet}, LaneNet~\cite{neven2018endtoend}, SCNN~\cite{pan2017spatial} and LDNet~\cite{munir2020ldnet}, are compared to our \textit{LaneSNNs} in \cref{tab:lane_comparison}.

\begin{table}[h]
    \centering
    \vspace*{4pt}
    \caption{Comparison of $IoU$ achieved by different algorithms~\cite{munir2020ldnet} for the lane detection problem faced by the semantic segmentation approach.}
    \label{tab:lane_comparison}
    \resizebox{\columnwidth}{!}{
    \begin{tabular}{c|c|c|c}
        Classifier &  $IoU_{offline}$ & $IoU_{online}$ & Number of parameters \\
        \hline
        FCN & $0.585$ & $-$ & $132.27$ M\\
        DeepLabv3 & $0.585$ & $-$ & $39.05$ M\\
        RefineNet & $0.614$ & $-$ & $99.02$ M\\
        LaneNet & $0.647$& $-$ & $0.53$ M\\
        SCNN & $0.673$& $-$ & $25.16$ M\\
        LDNet & $0.767$& $-$ & $5.71$ M\\
        \textit{CNN LaneSNN} (\textbf{ours}) & $0.598$& $0.349$ & $1.39$ M\\
        \textit{Fully-C600 LaneSNN} (\textbf{ours}) & $0.637$& $0.623$ & $1.20$ M\\
        \textit{Fully-C800 LaneSNN} (\textbf{ours}) & $0.633$& $0.613$ & $1.60$ M\\
        \textit{Fully-C800600 LaneSNN} (\textbf{ours}) & $0.652$& $0.550$ & $2.00$ M\\
    \end{tabular}
    }
    \vspace*{-15pt}
\end{table}

We can notice that our \textit{CNN LaneSNN} achieves higher performance than the FCN and DeepLabv3 algorithms despite they use more complex networks to make their predictions. The FCN uses AlexNet~\cite{alexNet} made of five convolutional layers, three pooling layers, and three fully-connected layers. DeepLabv3~\cite{chen2017deeplab} is made by Atrous Spatial Pyramid Pooling (ASPP) layers. It probes an incoming convolutional feature layer with filters at multiple sampling rates. This method is not yet developed onto the Loihi with only NxTF facilities~\cite{rueckauer2021nxtf} that do not implement the ASPP layers. Moreover, it has lower performance than RefineNet (based on the use of long residual connections), LaneNet (using upsampling layers and  conventional algorithms at the end of the process), SCNN (based on slice-by-slice convolutions), and LDNet (using ASPP, many convolution stacks, and upsampling layers). These structures cannot be developed onto the Loihi with the NxTF facilities~\cite{rueckauer2021nxtf} that implement pooling, fully connected, and traditional convolution layers. 

All the other fully-connected \textit{LaneSNNs} have $IoU$ comparable with LaneNet and overcome the performance of more complex algorithms. Moreover, the \textit{Fully-C800600 LaneSNN} achieves a similar result as the one obtained by the SCNN algorithm, while using less than $10\times$ number of parameters. On the other hand, the highest $IoU$ value has been measured by the LDNet, at the price of very high complexity and furthermore, it cannot be implemented onto the neuromorphic hardware.

The previous considerations do not take into account that all \textit{LaneSNNs} are tested on the modified DET dataset (discussed in \cref{cap_LaneSNN:Analysis_and_general_decisions:Dataset_pre-processing}) and not on the original, as it is for all the other presented algorithms. However, the DET dataset pre-processing step allows all the \textit{LaneSNNs} to be directly implementable on the Intel Loihi Neuromorphic Chip achieving competitive performance also online.

\section{Conclusion}
\label{sec:conclusion}

In this paper, we presented \textit{LaneSNNs}, a novel class of simple SNN models based on the \textit{semantic segmentation} approach, to find the position of the lanes on the driving road, thanks to event streams coming from an event-based camera. 
To the best of our knowledge, they represent the first SNNs implemented onto the Intel Loihi Neuromorphic Chip able to face the lane detection problem. 
Since they are implemented as an \textit{end-to-end} system, they do not use a post-processing stage of grouping and clustering. 
For training, we use a \textit{direct supervised} learning rule and we develop a \textit{novel loss function} that is the linear composition of WCE and MSE.

We design four structures with different complexity degrees and we call them \textit{CNN}, \textit{Fully-C600}, \textit{Fully-C800} and \textit{Fully-C800600}. The first is made of five convolution layers and a final fully-connected layer. The second and third are made by a fully-connected hidden layer made of 600 and 800 neurons respectively. The fourth is made of two fully connected hidden layers.


We train the \textit{LaneSNNs} with different parameters and then we implement the resulting SNNs onto the Intel Loihi Neuromorphic Chip. The best offline result for the offline implementation is achieved by the \textit{Fully-C800600 LaneSNN} with $IoU$ equal to $0.652$, while the best online result is achieved by the \textit{Fully-C600} network with $IoU$ equal to $0.623$. These values are comparable with the same achieved by other state-of-art algorithms such as LaneNet~\cite{neven2018endtoend} and RefineNet~\cite{lin2016refinenet}. Thanks to its implementation onto the Loihi Neuromorphic Research Chip, its maximum latency is less than $8\ ms$ and its power consumption is about $1\ W$ during the classification of a single image, thereby making it superior to all the state-of-the-art techniques in terms of performance and power efficiency.

\textit{LaneSNNs} enable high-performance yet energy-efficient lane detection for AD systems while leveraging the advanced neuromorphic processors.





\section*{Acknowledgments}

This work has been supported in part by the Doctoral College Resilient Embedded Systems, which is run jointly by the TU Wien’s Faculty of Informatics and the UAS Technikum Wien. This work was also supported in parts by the NYUAD’s Research Enhancement Fund (REF) Award on ``eDLAuto: An Automated Framework for Energy-Efficient Embedded Deep Learning in Autonomous Systems'', and by the NYUAD Center for Artificial Intelligence and Robotics (CAIR), funded by Tamkeen under the NYUAD Research Institute Award CG010.

\bibliographystyle{ieeetr}
\bibliography{main.bib}

\end{document}